\documentclass{article}
\pdfoutput=1
\usepackage[utf8]{inputenc}
\usepackage[a4paper,top=1in]{geometry}
\usepackage{url,microtype,subcaption}
\usepackage{hyperref}
\usepackage[onehalfspacing]{setspace}
\usepackage{amsmath}
\usepackage[font=small,labelfont=bf]{caption}
\usepackage{amsfonts}
\usepackage{wrapfig}
\usepackage{booktabs}
\usepackage{mathtools}
\usepackage{doi}
\usepackage[square,sort&compress,comma,numbers]{natbib}
\usepackage{amssymb}
\usepackage{authblk}
\usepackage{graphicx}
\usepackage[binary-units=true]{siunitx}
\usepackage{cleveref}
\usepackage{charter}
\usepackage{pifont,mdframed}

\newenvironment{warning}
  {\par\begin{mdframed}[linewidth=2pt,linecolor=red]%
    \begin{list}{}{\leftmargin=1cm
                   \labelwidth=\leftmargin}\item[\Large\ding{43}]}
  {\end{list}\end{mdframed}\par}

\renewenvironment{abstract}%
{%
\vskip 0.075in%
\centerline%
{\large\bf Abstract}%
\vspace{0.5ex}%
\begin{quote}%
}
{
\par%
\end{quote}%
\vskip 1ex%
}

\DeclareMathOperator*{\argmax}{\arg\!\max}
\DeclarePairedDelimiter\abs{\lvert}{\rvert}%
\author[1,*]{Timo Wunderlich}
\author[1,**]{Akos F. Kungl}
\author[1]{Eric Müller}
\author[1]{Andreas Hartel}
\author[1]{Yannik Stradmann}
\author[1]{Syed Ahmed Aamir}
\author[1]{Andreas Grübl}
\author[1]{Arthur Heimbrecht}
\author[1]{Korbinian Schreiber}
\author[1]{David Stöckel}
\author[1]{Christian Pehle}
\author[1]{Sebastian Billaudelle}
\author[1]{Gerd Kiene}
\author[1]{Christian Mauch}
\author[1]{Johannes Schemmel}
\author[1]{Karlheinz Meier}
\author[1,2]{Mihai A. Petrovici}
\affil[1]{Kirchhoff Institute for Physics, Heidelberg University, Heidelberg, Germany}
\affil[2]{Department of Physiology, University of Bern, Bern, Switzerland}
\affil[*]{timo.wunderlich@kip.uni-heidelberg.de}
\affil[**]{fkungl@kip.uni-heidelberg.de}

\date{}
\begin{document}

\title{Demonstrating Advantages of Neuromorphic Computation: A Pilot Study}
\begin{warning}
This article is bound to be published in Frontiers in Neuromorphic Engineering (\url{https://www.frontiersin.org/articles/10.3389/fnins.2019.00260}).\\
Please cite it as \emph{Wunderlich et al., Frontiers in Neuromorphic Engineering (2019), doi: 10.3389/fnins.2019.00260}.
\end{warning}
{\let\newpage\relax\maketitle}

\begin{abstract}
\begin{singlespace}
Neuromorphic devices represent an attempt to mimic aspects of the brain's architecture and dynamics with the aim of replicating its hallmark functional capabilities in terms of computational power, robust learning and energy efficiency.
We employ a single-chip prototype of the BrainScaleS~2 neuromorphic system to implement a proof-of-concept demonstration of reward-modulated spike-timing-dependent plasticity in a spiking network that learns to play a simplified version of the Pong video game by smooth pursuit.
This system combines an electronic mixed-signal substrate for emulating neuron and synapse dynamics with an embedded digital processor for on-chip learning, which in this work also serves to simulate the virtual environment and learning agent.
The analog emulation of neuronal membrane dynamics enables a 1000-fold acceleration with respect to biological real-time, with the entire chip operating on a power budget of \SI{57}{\milli\watt}.
Compared to an equivalent simulation using state-of-the-art software, the on-chip emulation is at least one order of magnitude faster and three orders of magnitude more energy-efficient.
We demonstrate how on-chip learning can mitigate the effects of fixed-pattern noise, which is unavoidable in analog substrates, while making use of temporal variability for action exploration.
Learning compensates imperfections of the physical substrate, as manifested in neuronal parameter variability, by adapting synaptic weights to match respective excitability of individual neurons.

\end{singlespace}
\end{abstract}
\section{Introduction}
Neuromorphic computing represents a novel paradigm for non-Turing computation that aims to reproduce aspects of the ongoing dynamics and computational functionality found in biological brains.
This endeavor entails an abstraction of the brain's neural architecture that retains an amount of biological fidelity sufficient to reproduce its functionality while disregarding unnecessary detail.
Models of neurons, which are considered the computational unit of the brain, can be emulated using electronic circuits or simulated using specialized digital systems \citep{Indiveri2011NeuromorphicCircuits,furber2016_largescale}.

BrainScaleS~2 (BSS2) is a neuromorphic architecture consisting of CMOS-based ASICs \citep{Aamir2018AnArchitecture,Friedmann2017DemonstratingSystem} which implement physical models of neurons and synapses in analog electronic circuits while providing facilities for user-defined learning rules.
A number of features distinguish BSS2 from other neuromorphic approaches, such as a speed-up factor of $10^3$ compared to biological neuronal dynamics, correlation sensors for spike-timing-dependent plasticity in each synapse circuit and an embedded processor \citep{Friedmann2017DemonstratingSystem}, which can use neural network observables to calculate synaptic weight updates for a broad range of plasticity rules.
The flexibility enabled by the embedded processor is a particularly useful feature given the increasing effort invested in synaptic plasticity research, allowing future findings to be accommodated easily.
The study at hand uses a single-chip prototype version of the full system, which allows the evaluation of the planned system design on a smaller scale.

Reinforcement learning has been prominently used as the learning paradigm of choice in machine learning systems which reproduced, or even surpassed, human performance in video and board games \citep{Mnih2015Human-levelLearning,Silver2016MasteringSearch,Silver2017MasteringKnowledge}.
In reinforcement learning, an agent interacts with its environment and receives reward based on its behavior.
This enables the agent to adapt its internal parameters so as to increase the potential for reward in the future \citep{Sutton1998ReinforcementIntroduction}.
In the last decades, research has found a link between reinforcement learning paradigms used in machine learning and reinforcement learning in the brain \citep[for a review, see][]{Niv2009ReinforcementBrain}. 
The neuromodulator dopamine was found to convey a reward prediction error, akin to the temporal difference error used in  reinforcement learning methods \citep{Schultz1997AReward.,Sutton1998ReinforcementIntroduction}.
Neuromodulated plasticity can be modeled using three-factor learning rules \citep{Fremaux2015NeuromodulatedRules.,Fremaux2013ReinforcementNeurons}, where the synaptic weight update depends not only on the learning rate and the pre- and post-synaptic activity but also on a third factor, representing the neuromodulatory signal, which can be a function of reward, enabling reinforcement learning.

In this work, we demonstrate the advantages of neuromorphic computation by showing how an agent controlled by a spiking neural network (SNN) learns to solve a smooth pursuit task via reinforcement learning in a fully embedded perception-action loop that simulates the classic Pong video game on the BSS2 prototype.
Measurements of time-to-convergence, power consumption and sensitivity to parameter noise demonstrate the  advantages of our neuromorphic solution compared to classical simulation on a modern CPU that runs the NEST simulator \citep{Peyser2017NEST2.14.0}.
The on-chip learning converges within seconds, which is equivalent to hours in biological terms, while the software simulation is at least an order of magnitude slower and three orders of magnitude less energy-efficient.
We find that fixed-pattern noise on BSS2 can be compensated by the chosen learning paradigm, reducing the required calibration precision, and that the results of hyperparameter learning can be transferred between different BSS2 chips.
The experiment takes place on the chip fully autonomously, i.e., both the environment and synaptic weight changes are computed using the embedded processor.
As the number of neurons (32) and synapses (1024) on the prototype chip constrain the complexity of solvable learning tasks, the agent's task in this work is simple smooth pursuit without anticipation.
The full system is expected to enable more sophisticated learning, akin to the learning of Pong from pixels that was previously demonstrated using an artificial neural network \citep{Mnih2015Human-levelLearning}.

\section{Materials and Methods}
\label{sec:mandm}

\subsection{The BrainScaleS~2 Neuromorphic Prototype Chip}\label{sec:dls}

The BSS2 prototype is a neuromorphic chip and the predecessor of a large-scale accelerated network emulation platform with flexible plasticity rules \citep{Friedmann2017DemonstratingSystem}.
It is manufactured using a \SI{65}{\nano\meter} CMOS process and is designed for mixed-signal neuromorphic computation.
All experiments in this work were performed on the second prototype version.
Future chips will be integrated into a larger setup using wafer-scale technology  \citep{Schemmel2010AModeling,Zoschke2017FullCluster}, thereby enabling the emulation of large plastic neural networks.

\begin{figure}[h]
\begin{center}
\includegraphics[width=15cm]{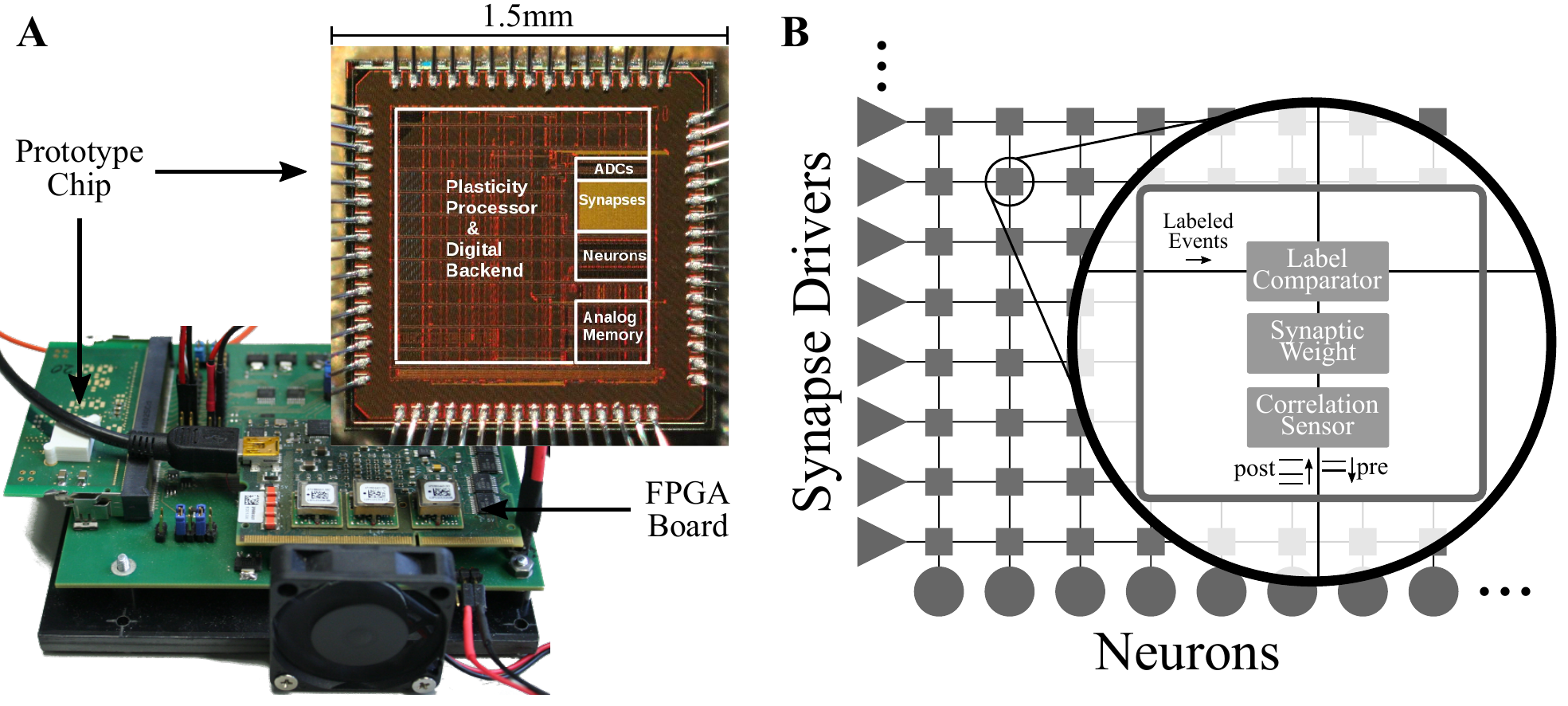}
\end{center}
\caption{\label{fig:dls}%
Physical setup and neural network schematic. 
\textbf{A}: In the foreground: BSS2 prototype chip with demarcation of different functional parts.
In the background: the development board on which the chip is mounted. 
Adapted from \citep{Aamir2018AnArchitecture}.
\textbf{B}: Schematic of the on-chip neural infrastructure.
Each of the $32$ implemented neurons is connected to one column of the synapse array, where each column comprises $32$ synapses.
Synapse drivers allow row-wise injection of individually labeled (6-bit) spike events.
Each synapse locally stores a 6-bit label and a 6-bit weight and converts spike events with a matching label to current pulses traveling down towards the neuron. 
Each synapse also contains an analog sensor measuring the temporal correlation of pre- and post-synaptic events (see \Cref{sec:stdp}).
}
\end{figure}

\subsubsection{Experimental Setup}\label{sec:hwandswenv}

The BSS2 prototype setup is shown in \Cref{fig:dls} A and contains the neuromorphic chip mounted on a prototyping board.
The chip and all of its functional units can be accessed and configured from either a Xilinx Spartan-6 FPGA or the embedded processor (see \Cref{sec:ppu}).
The FPGA in turn can be accessed via a USB-2.0 connection between the prototype setup and the host computer.
In addition to performing chip configuration, the FPGA can also provide hard real-time playback of input and recording of output data.

Experiments are described by the user through a container-based programming interface which provides access to all functional units such as individual neuron circuits or groups of synapses.
The experiment configuration is transformed into a bitstream and uploaded to DRAM attached to the FPGA.
Subsequently, the software starts the experiment and a sequencer logic in the FPGA begins to play back the experiment data (e.g., input spike trains) stored in the DRAM.
At the same time, output from the chip is recorded to a different memory area in the DRAM.
Upon completion of the experiment, the host computer downloads all recorded output from the FPGA memory.

\subsubsection{Neurons \& Synapses}\label{sec:neuronsandsynapses}

Our approach to neuromorphic engineering follows the idea of ``physical modeling'': the analog neuronal circuits are designed to have similar dynamics compared to their biological counterparts, making use of the physical characteristics of the underlying substrate.
The BSS2 prototype chip contains $32$ analog neurons based on the Leaky Integrate-and-Fire (LIF) model \citep{Aamir2018AnArchitecture,Aamir2016ASystem}.
Additionally, each neuron has an $8$-bit spike counter, which can be accessed and reset by the the embedded processor (\cite{Friedmann2017DemonstratingSystem}, see \Cref{sec:ppu}) for plasticity-related calculations.

In contrast to other neuromorphic approaches \citep{Qiao2015ASynapses, Furber2014TheProject,Benjamin2014Neurogrid:Simulations,Merolla2014AInterface,Davies2018Loihi:Learning}, this implementation uses the fast supra-threshold dynamics of CMOS transistors in circuits which mimic neuronal membrane dynamics.
In the case of BSS2, this approach provides time constants that are smaller than their biological counterparts by three orders of magnitude, i.e.,  the hardware operates with a speed-up factor of \num{e3} compared to biology, independent of the network size or plasticity model.
Throughout the manuscript, we provide the true (wall-clock time) values, which are typically on the order of microseconds, compared to the millisecond-scale values usually found in biology.

The $32$-by-$32$ array of synapses is arranged such that each neuron can receive input from a column of $32$ synapses (see \Cref{fig:dls} B).
Each row consisting of 32 synapses can be individually configured as excitatory or inhibitory and receives input from a synapse driver that injects labeled digital pre-synaptic spike packets. 
Every synapse compares its label (a locally stored configurable address) with the label of a given spike packet and if they match, generates a current pulse with an amplitude proportional to its $6$-bit weight that is sent down along the column towards the post-synaptic neuron.
There, the neuron circuit converts it into an exponential post-synaptic current (PSC), which is injected into the neuronal membrane capacitor.

Post-synaptic spikes emitted by a neuron are signaled (back-propagated) to every synapse in its column, which allows the correlation sensor in each synapse to record the time elapsed between pre- and post-synaptic spikes.
Thus, each synapse accumulates correlation measurements that can be read out by the embedded processor, to be used, among other observables, for calculating weight updates (see \Cref{sec:stdp} for a detailed description).

\subsubsection{Calibration and Configuration of the Analog Neurons}
\label{sec:calib}

Neurons are configured using on-chip analog capacitive memory cells \citep{Hock2013AnHardware}.
The ideal LIF model neuron with one synapse type and exponential PSCs can be characterized by six parameters: membrane time constant $\tau_\textrm{mem}$, synaptic time constant $\tau_\textrm{syn}$, refractory period $\tau_\textrm{ref}$, resting potential $v_\textrm{leak}$, threshold potential $v_\textrm{thresh}$, reset potential $v_\textrm{reset}$.
The neuromorphic implementation on the chip carries $18$ tunable parameters per neuron and one global parameter \citep{Aamir2018AnArchitecture}.
Most of these hardware parameters are used to set the circuits to the proper point of operation and therefore have fixed values that are calibrated once for any given chip; for the experiments described here, the six LIF model parameters mentioned above are fully controlled by setting only six of the hardware parameters per neuron.

Manufacturing variations cause fixed-pattern noise (see \Cref{sec:noise}), therefore each neuron circuit behaves differently for any given set of hardware parameters.
In particular, the time constants ($\tau_\textrm{mem}$, $\tau_\textrm{syn}$, $\tau_\textrm{ref}$) display a high degree of variability.
Therefore, in order to accurately map user-defined LIF time constants to hardware parameters, neuron circuits are calibrated individually. Using this calibration data reduces deviations from target values to less than \SI{5}{\percent} \citep[][see also \Cref{fig:trial-to-trial} B]{Aamir2018AnArchitecture}.

\subsubsection{Plasticity Processing Unit}\label{sec:ppu}

To allow for flexible implementation of plasticity algorithms, the chip uses a Plasticity Processing Unit (PPU), which is a general-purpose $32$-bit processor implementing the PowerPC-ISA 2.06 instruction set and custom vector extensions \citep{Friedmann2017DemonstratingSystem}.
In the used prototype chip, it is clocked at a frequency of \SI{98}{\mega\hertz} and has access to \SI{16}{\kibi\byte} of main memory.
Vector registers are $128$-bit wide and can be processed in slices of eight $16$-bit or sixteen $8$-bit units within one clock cycle.
The vector extension unit is loosely coupled to the general-purpose part.
When fetching vector instructions, the commands are inserted into a dedicated command queue which is read by the vector unit.
Vector commands are decoded, distributed to the arithmetic units and executed as fast as possible.

The PPU has dedicated fully-parallel access ports to synapse rows, enabling row-wise readout and configuration of synaptic weights and labels.
This enables efficient row-wise vectorized plasticity processing.
Modifications of connectivity, neuron and synapse parameters are supported during neural network operation.
The PPU can be programmed using assembly and higher-level languages such as \textit{C} or \textit{C}\texttt{++} to compute a wide range of plasticity rules.
Compiler support for the PPU is provided by a customized \texttt{gcc} \citep{github:electronicvisions:gcc,stallman2018gcc}.
The software used in this work is written in \textit{C}, with vectorized plasticity processing implemented using inline assembly instructions.

\subsubsection{Correlation Measurement at the Synapses}
\label{sec:stdp}
\begin{figure}[h]
\begin{center}
\includegraphics[width=15cm]{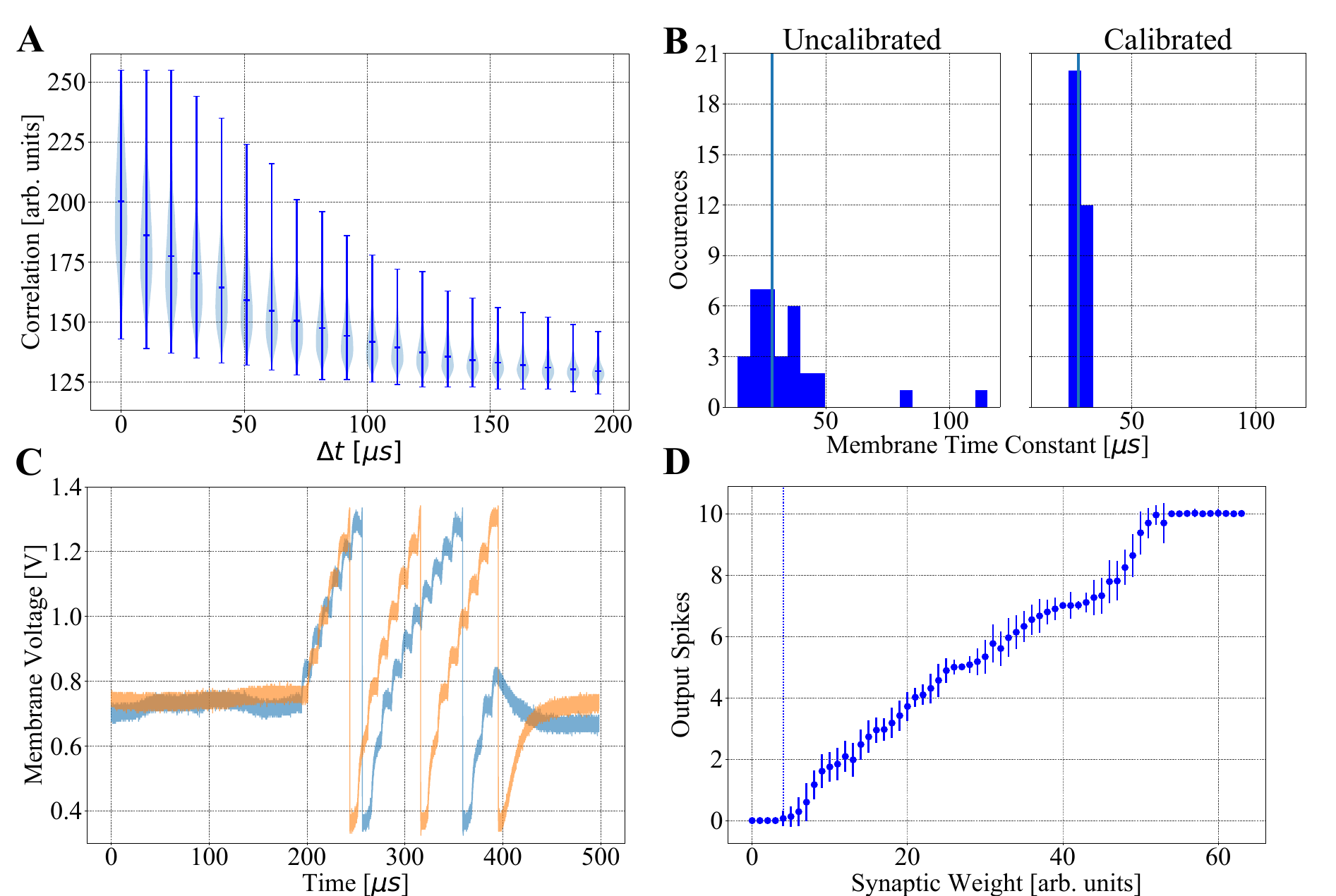}
\caption{\label{fig:trial-to-trial}%
BSS2 is subject to fixed-pattern noise and temporal variability.
\textbf{A}: Violin plot of the digitized output of the  $1024$ causal correlation sensors ($a_+$, see \Cref{eq:causal-correlation}) on a sample chip (chip \#1) as a function of the time interval between a single pre-post spike pair.
\textbf{B}: Distribution of membrane time constants $\tau_\mathrm{m}$ over all 32 neurons with and without calibration.
The target value is \SI{28.5}{\micro\second} (vertical blue lines).
\textbf{C}: Effects of temporal variability.
A regular input spike train containing twenty spikes spaced by \SI{10}{\micro\second}, as used in the learning task, transmitted via one synapse, elicits different membrane responses in two trials.
\textbf{D}: Mean and variance of the output spike count as a function of synaptic weight, averaged over $100$ trials, for a single exemplary neuron receiving the input spike train from (C). 
The spiking threshold weight (the smallest weight with a higher than \SI{5}{\percent} probability of eliciting an output spike under the given stimulation paradigm) is indicated by the dotted blue line.
Trial-to-trial variation of the number of output spikes at fixed synaptic weight is due to temporal variability and mediates action exploration. 
}
\end{center}
\end{figure}

Every synapse in the synapse array contains two analog units that record the temporal correlation between nearest-neighbor pairs of pre- and post-synaptic spikes.
For each such pair, a dedicated circuit measures the value of either the causal (pre before post) or anti-causal (post before pre) correlation, which is modeled as an exponentially decaying function of the spike time difference \citep{Friedmann2017DemonstratingSystem}.
The values thus measured are accumulated onto two separate storage capacitors per synapse.
In an idealized model, the voltages across the causal and anti-causal storage capacitor are 
\begin{equation} \label{eq:causal-correlation}
a_+ = \sum_\mathrm{pre-post} \eta_+ \exp\left( -\frac{t_\mathrm{post} - t_\mathrm{pre}}{\tau_+} \right)
\end{equation}
and
\begin{equation}
a_- = \sum_\mathrm{post-pre} \eta_- \exp\left( -\frac{t_\mathrm{pre} - t_\mathrm{post}}{\tau_-} \right) \; ,
\end{equation}
respectively, with decay time constants $\tau_+$ and $\tau_-$ and scaling factors $\eta_+$ and $\eta_-$.
These accumulated voltages represent non-decaying eligibility traces that can be read out by the PPU using column-wise $8$-bit Analog-to-Digital Converters (ADCs), allowing row-wise parallel readout.
Fixed-pattern noise introduces variability among the correlation units of different synapses, as visible in \Cref{fig:trial-to-trial} A.
The experiments described here only use the causal traces $a_+$ to calculate weight updates.

\subsection{Types of Noise on BSS2}
\label{sec:noise}

The BSS2 prototype has several sources of parameter variability and noise, as does any analog hardware.
We distinguish between fixed-pattern noise and temporal variability.

\emph{Fixed-pattern noise} refers to the systematic deviation of  parameters (e.g., transistor parameters) from the values targeted during chip design.
This type of noise is caused by the inaccuracies of the manufacturing process and unavoidable stochastic variations of process parameters.
Fixed-pattern noise is constant in time and induces heterogeneity between neurons and synapses, but  calibration can reduce it to some degree.
The effects of the calibration on the distribution of the membrane time constant $\tau_\textrm{mem}$ are shown in \Cref{fig:trial-to-trial} B.

\emph{Temporal variability} continually influences circuits during their operation, leading to fluctuations of important dynamical variables such as membrane potentials. 
Typical sources of temporal variability are crosstalk, thermal noise and the limited stability of the analog parameter storage.
These effects can cover multiple timescales and lead to variable neuron spike responses, even when the input spike train remains unchanged between trials.
A concrete example of trial-to-trial variability of a neuron's membrane potential evolution, output spike timing and firing rate caused by temporal variability is shown in \Cref{fig:trial-to-trial} C, D for two trials of the same experiment, using the same input spike train and parameters, with no chip reconfiguration between trials.

\subsection{Reinforcement Learning with Reward-Modulated STDP}

In reinforcement learning, a behaving agent interacts with its environment and tries to maximize the expected future reward it receives from the environment as a consequence of this interaction \citep{Sutton1998ReinforcementIntroduction}.
The techniques developed to solve problems of reinforcement learning generally do not involve spiking neurons and are not designed to be biologically plausible. 
Yet reinforcement learning evidently takes place in biological SNNs, e.g., in basic operant conditioning \citep{Guttman1953OperantAgent.,Moritz2011VolitionalInterface.,Fetz1973OperantlyMuscles.}. The investigation of spike-based implementations with biologically inspired plasticity rules is therefore an interesting subject of research with evident applications for neuromorphic devices.
The learning rule used in this work, Reward-modulated Spike-Timing Dependent Plasticity (R-STDP) \citep{Izhikevich2007SolvingSignaling,Farries2007ReinforcementPlasticity,Fremaux2010FunctionalPlasticity}, represents one possible implementation.

R-STDP is a three-factor learning rule that modulates the effect of unsupervised STDP using a reward signal.
Recent work has used R-STDP to reproduce Pavlovian conditioning as in \cite{Izhikevich2007SolvingSignaling} on a specialized neuromorphic digital simulator, achieving real-time simulation speed \citep{mikaitis2018_neuromodulated}.
While not yet directly applied to an analog neuromorphic substrate, aspects of this learning paradigm have already been studied in software simulations, under constraints imposed by the BSS2 system, in particular concerning the effect of discretized weights, with promising results \citep{Friedmann2013Reward-basedSubstrate}.
Furthermore, it was previously suggested that trial-to-trial variations of neuronal firing rates as observed in cortical neurons can benefit learning in a reinforcement learning paradigm, rather than being a nuisance \citep{Legenstein2008ABiofeedback,Maass2014NoiseNeuronsb,Xie2004LearningSpiking}.
Our experiments corroborate this hypothesis by explicitly using trial-to-trial variations due to temporal variability for action exploration.

The reward mechanism in R-STDP is biologically inspired: the phasic activity of dopamine neurons in the brain was found to encode expected reward \citep{Hollerman1998DopamineLearning,Bayer2005MidbrainSignal,Schultz1997AReward.} and dopamine concentration modulates STDP \citep{Pawlak2008Cellular/MolecularPlasticity,Brzosko2015RetroactiveDopamine,Edelmann2011DopamineSlices}.
R-STDP and similar reward-modulated Hebbian learning rules have been used to solve a variety of learning tasks in simulations, such as reproducing temporal spike patterns and spatio-temporal trajectories \citep{Fremaux2010FunctionalPlasticity,Farries2007ReinforcementPlasticity,Vasilaki2009Spike-BasedFail}, reproducing the results of classical conditioning \citep{Izhikevich2007SolvingSignaling}, making a recurrent neural network exhibit specific periodic activity and working-memory properties \citep{Hoerzer2014EmergenceLearning} and reproducing the seminal biofeedback experiment by Fetz and Baker \citep{Legenstein2008ABiofeedback,Fetz1973OperantlyMuscles.}.
Compared to classic unsupervised STDP, using R-STDP was shown to improve the performance of a spiking convolutional neural network tasked with visual categorization \citep{Mozafari2018First-Spike-BasedSTDP,Mozafari2018CombiningRecognition}. 

In contrast to other learning rules in reinforcement learning, R-STDP is not derived using gradient descent on a loss function; rather, it is motivated heuristically \citep{Fremaux2015NeuromodulatedRules.}, the idea being to multiplicatively modulate STDP using a reward term. 

We employ the following form of discrete weight updates using R-STDP:
\begin{equation}
\Delta w_{ij} = \beta \cdot (R-b)\cdot e_{ij} \; ,
\end{equation}
where $\beta$ is the learning rate, $R$ is the reward, $b$ is a baseline and $e_{ij}$ is the STDP eligibility trace  which is a function of the pre- and post-synaptic spikes of the synapse connecting neurons $i$ and $j$.
The choice of the baseline reward $b$ is critical: a nonzero offset introduces an admixture of unsupervised learning via the unmodulated $\textrm{STDP}$ term, and choosing $b$ to be the task-specific expected reward $b=\langle R\rangle_\textrm{task}$ leads to weight updates that capture the covariance of reward and synaptic activity \citep{Fremaux2015NeuromodulatedRules.}:
\begin{equation}
\langle \Delta w_{ij} \rangle_\textrm{task} = \langle R \cdot e_{ij}\rangle_\textrm{task} - \langle R  \rangle_\textrm{task} \cdot \langle e_{ij} \rangle_\textrm{task} = \mathrm{Cov}(R, e_{ij}) \; .
\end{equation}

This setting, which we also employ in our experiments, makes R-STDP a statistical learning rule in the sense that it captures correlations of joint pre- and post-synaptic activity and reward; this information is collected over many trials of any single learning task.
The expected reward may be estimated as a moving average of the reward over the last trials of that specific task; task specificity of the expected reward is required when multiple tasks need to be learned in parallel \citep{Fremaux2010FunctionalPlasticity}.

\begin{figure}[h]
\begin{center}
\includegraphics[width=15cm]{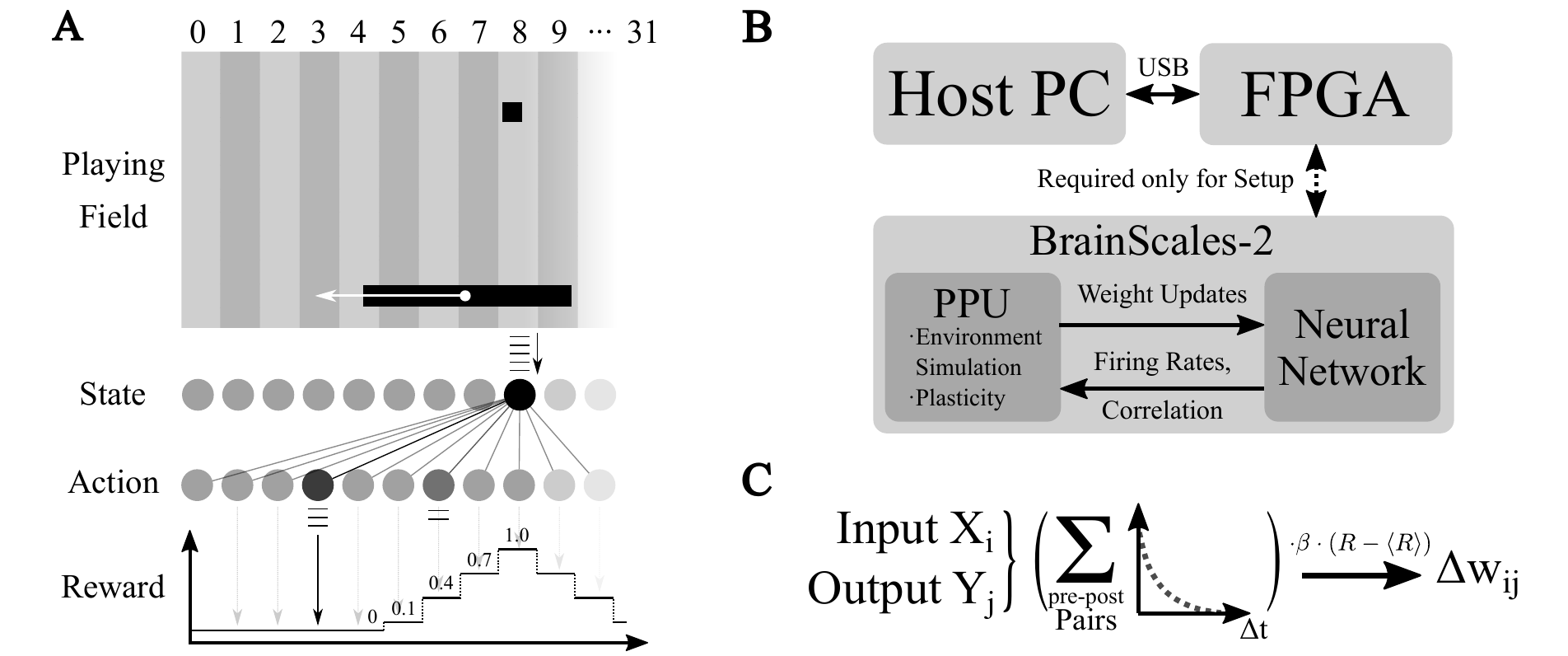}
\end{center}
\caption{\label{fig:exp-setup}%
Overview of the experimental setup.
\textbf{A}: The components of the environment are the playing field with three reflective walls (top, left, right), the ball and the paddle.
In the depicted situation, the ball is in column $8$ and therefore a uniform spike train is sent to output neurons via the synapses of input unit $8$.
Output unit $3$ fires the most spikes and the paddle therefore moves toward column $3$.
As the target column is too far away from the ball column, it lies outside of the graded reward window and the reward received by the network is zero (see \Cref{eq:reward}). 
\textbf{B}: The chip performs the experiment completely autonomously: the environment simulation, spiking network emulation and synaptic plasticity via the PPU are all handled on-chip.
The FPGA, which is only used for the initial configuration of the chip, is controlled via USB from the host PC. 
\textbf{C}: The plasticity rule calculated by the PPU.
Pre-before-post spike pairs of input unit $X_i$ and output unit $Y_j$ are exponentially weighted with the corresponding temporal distance and added up.
This correlation measure is then multiplied with the learning rate and the difference of instantaneous and expected reward to yield the weight change for synapse $(i, j)$.}
\end{figure}

\subsection{Learning Task and Simulated Environment}

\begin{figure}[h]
    \begin{center}
        \includegraphics[width=10cm]{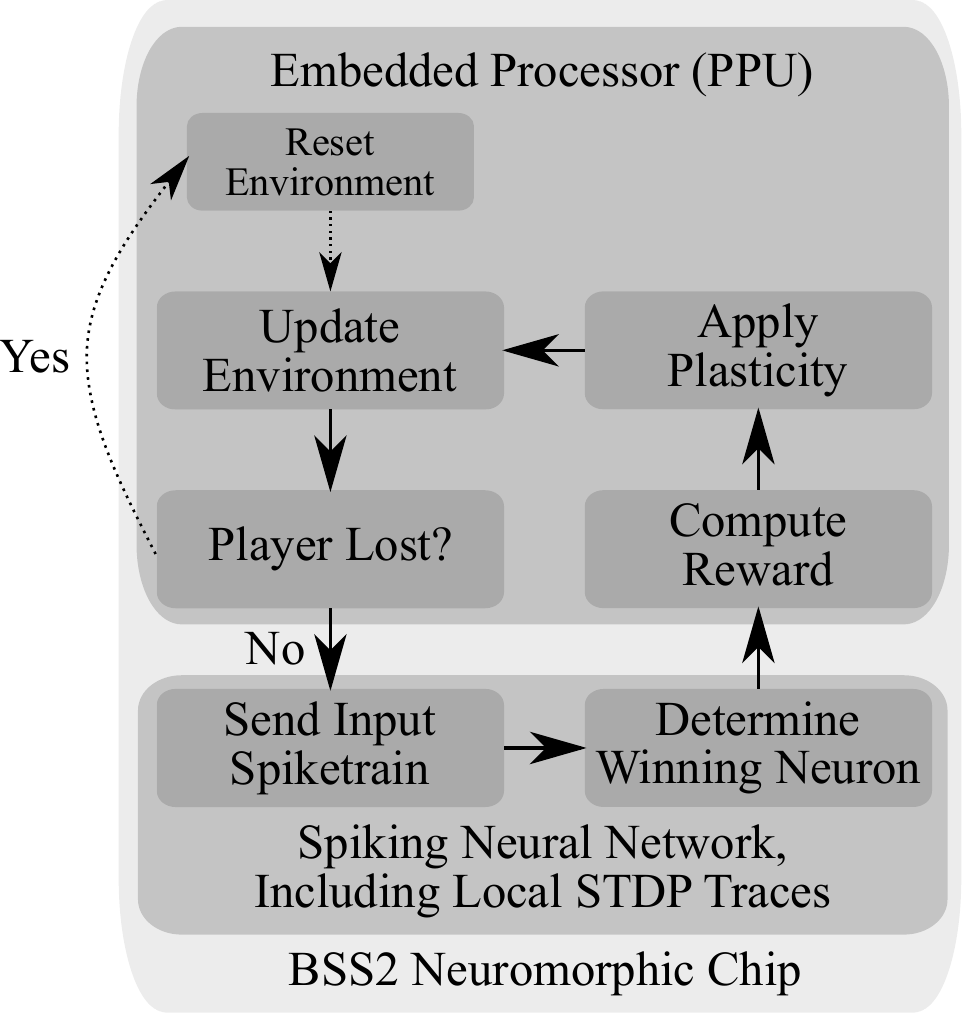}
    \end{center}
    \caption{Flowchart of the experiment loop running autonomously on BSS2, using both the analog Spiking Neural Network (SNN) and the embedded processor. 
    The environment is reset by positioning the ball in the middle of the playing field with a random direction of movement at the start of the experiment or upon the agent's failure to reflect the ball. 
    In the main loop, the (virtual) state unit corresponding to the current ball position transmits a spike train to all neurons in the action layer (see \Cref{fig:exp-setup}).
    Afterwards, the winning action neuron, i.e., the action neuron that had the highest output spike count, is determined.
    Then, the reward is determined based on the difference between the ball position and the target paddle position as dictated by the winning neuron (\Cref{eq:reward}).
    Using the reward, a stored running average of the reward and the STDP correlation traces computed locally at each synapse during SNN emulation, the weight updates of all synapses are computed (\Cref{eq:deltawmn}) and applied.
    The environment, i.e., the position of ball and paddle, are then updated, and the loop starts over.}
    \label{fig:flowchart}
\end{figure}

Using the PPU, we simulate a simple virtual environment inspired by the Pong video game.
The components of the game are a two-dimensional square playing field, a ball and the player's paddle (see \Cref{fig:exp-setup} A).
The paddle is controlled by the chip and the goal of the learning task is to trace the ball.
Three of the four playing field sides are solid walls, with the paddle moving along the open side.
The experiment proceeds iteratively and fully on-chip (see \Cref{fig:exp-setup} B).
A single experiment iteration consists of neural network emulation, weight modification and environment simulation.
We visualize one iteration as a flowchart in \Cref{fig:flowchart} and provide a detailed account in the following.

The game dynamics consist of the ball starting in the middle of the playing field in a random direction with velocity $\vec v$ and the paddle which moves with velocity $v_\mathrm{p}$ along its baseline. 
Surfaces elastically reflect the ball.
If the paddle misses the ball, the game is reset, i.e., the ball starts from the middle of the field in a random direction.
Specific parameter values are given in \Cref{tab:parameters}.

The paddle is controlled by the neural network emulated on the neuromorphic chip.
The network receives the ball position along the paddle's baseline as input and determines the movement of its paddle via the resulting network activity.
The neural network consists of two $32$-unit layers which are initialized with all-to-all feed-forward connections (see \Cref{tab:parameters}), where the individual weights are drawn from a Gaussian distribution and where the first layer represents input/state units $u_i$ and the second layer represents output/action units $v_i$.
The first layer is a virtual layer for providing input spikes and the second layer consists of the chip's LIF neurons. 
All synaptic connections are excitatory.
Discretizing the playing field into $32$ columns along the paddle baseline, we assign a state unit $u_i$ to each column $i$ where $i\in[0,31]$ and provide input to this unit if the ball is in the corresponding column via a uniform spike train (see \Cref{tab:parameters} for parameters).

The action neurons' spike counts $\rho_i$ are used to determine the paddle movement: the unit with the highest number of output spikes $j=\argmax_i (\rho_i)$ determines the paddle's target column $j$, towards which it moves with constant velocity $v_\mathrm{p}$ (see \Cref{tab:parameters}).
If the target column and center position of the paddle match, no movement is performed.
Spike counts and in consequence, the target row $j$ are determined after the input spike train has been delivered.
If several output units have the same spike count, the winner is chosen randomly among them.
Afterwards, the reward $R$ is calculated based on the distance between the target column $j$ and the current column of the ball, $k$: 
\begin{equation}
R = \begin{cases}
 1 - \abs{j-k}\cdot 0.3 &\text{ if } \abs{j-k}\leq 3 \; , \\
 0 &\text{ otherwise.}
 \end{cases} \label{eq:reward}
\end{equation} 
Learning success is therefore graded, i.e., the network obtains reduced reward for less than optimal aiming.
The size of the reward window defined by \Cref{eq:reward} is chosen to match the paddle length.

For every possible task (corresponding to a ball column $k$), the PPU holds task-specific expected rewards $\bar R_k$, $k\in[0,31]$ in its memory, which it subtracts from the instantaneous reward $R$ to yield the neuromodulating factor $R-\bar R_k$, which is also used to update the task-specific expected reward as an exponentially weighted moving average:
\begin{equation}
\bar R_k \leftarrow \bar R_k + \gamma \left(R-\bar R_k\right) \; , 
\end{equation} 
where $\gamma$ controls the influence of previous iterations.
The expected reward of any state is initialized as the first reward received in that state.
All $1024$ synapse weights $w_{mn}$ from input unit $m$ to output unit $n$ are then updated according to the three-factor rule (see \Cref{fig:exp-setup} C)
\begin{equation}
\Delta w_{mn} = \beta \cdot \left(R-\bar R_k\right) \cdot A^+_{mn} \; ,
\label{eq:deltawmn}
\end{equation} 
where $\beta$ is a learning rate and $A^+_{mn}$ is a modified version of $a^+_{mn}$ (see \Cref{eq:causal-correlation}) which has been corrected for offset, digitized to 8 bit and right-shifted by one bit in order to reduce noise.
After the weights have been modified, the Pong environment is updated according to the described dynamics and the next iteration begins.

The \emph{mean expected reward}
\begin{equation}
 \langle \bar R \rangle = \frac{1}{32} \sum_{i=0}^{31} \bar R_i \; , \label{eq:mean-reward}
\end{equation}
i.e., the average of the expected rewards over all states, represents a measure of the progress of learning in any given iteration.
Due to the paddle width and correspondingly graded reward scheme, the agent is able to catch the ball even when it is not perfectly centered below the ball.
Therefore, the \emph{performance} in playing Pong can be quantified by
\begin{equation}
    P = \frac{1}{32} \sum_{i=0}^{31}  \lceil R_i \rceil \; , \label{eq:performance}
\end{equation}
where $R_i$ is the last reward received in state $i$. 
This provides the percentage of states in which the agent has aimed the paddle such that it is able to catch the ball.

In order to find suitable hyperparameters for the chip configuration, we used a Bayesian parameter optimization based on sequential optimization using decision trees to explore the neuronal and synaptic parameter space while keeping parameters  such as game dynamics, input spike train and initial weight distribution fixed. 
The software package used was scikit-optimize \citep{Head2018Scikit-optimize/scikit-optimize:V0.5.2}.
Initially, 30 random data points were taken, followed by 300 iterations with the next evaluated parameter set being determined by the optimization algorithm \textsc{forest\_minimize} with default parameters (maximizing expected improvement, extra trees regressor model, 10000 acquisition function samples, target improvement of 0.01). 
All neuron and synapse parameters were subject to the optimization, i.e., all neuronal time constants and voltages as well as the time constant and amplitude of the synapse correlation sensors.
The results are a common set of parameters for all neurons and synapses (see \Cref{tab:parameters}).

\subsection{Software Simulation with NEST}
\label{sec:sim}

In order to compare the learning performance and speed of the chip to a network of deterministic, perfectly identical LIF neurons, we ran a software simulation of the experiment using the NEST v2.14.0 SNN simulator \citep{Peyser2017NEST2.14.0}, using the same target LIF parameters as in our experiments using the chip, with time constants scaled by a factor of \num{e3}.
We did not include fixed-pattern noise of neuron parameters in the simulation, i.e., all neurons had identical parameters.
We used the \textit{iaf\_psc\_exp} integrate-and-fire neuron model available in NEST, with exponential PSC kernels and current-based synapses.
Using NEST's \textit{noise\_generator}, we are able to investigate the effect of injecting Gaussian current noise into each neuron.
The scaling factors $\eta_+$ and $\eta_-$, as well as the time constants $\tau_+$ and $\tau_-$ of the correlation sensors were chosen to match the mean values on BSS2.
The correlation factor $a_+$ was calculated within the Python script controlling the experiment using \Cref{eq:causal-correlation} and the spike times provided by the NEST simulator.
Hyperparameters such as learning rate and game dynamics (e.g. the reward window defined in \Cref{eq:reward}) were set to be equivalent to BSS2 and weights were scaled to a dimensionless quantity and discretized to match the neuromorphic emulation.

The synaptic weight updates in each iteration were restricted to those synapses which transmitted spikes, i.e., the synapses from the active input unit to all output units ($32$ out of the $1024$ synapses), as the correlation $a_+$ of all other synapses is zero in a perfect simulation without fixed-pattern noise.
This has the effect of reducing the overall time required to simulate one iteration and is in contrast to the implementation on BSS2, where all synapses are updated in each iteration as there is no guarantee that correlation traces are zero and we excluded this kind of ``expert knowledge'' from the implementation. 

The source code of the simulation is publicly available \citep{Wunderlich2019NeuromorphicSimulation}.

\section{Results}
\begin{figure}[h]
    \centering
    \includegraphics[width=15cm]{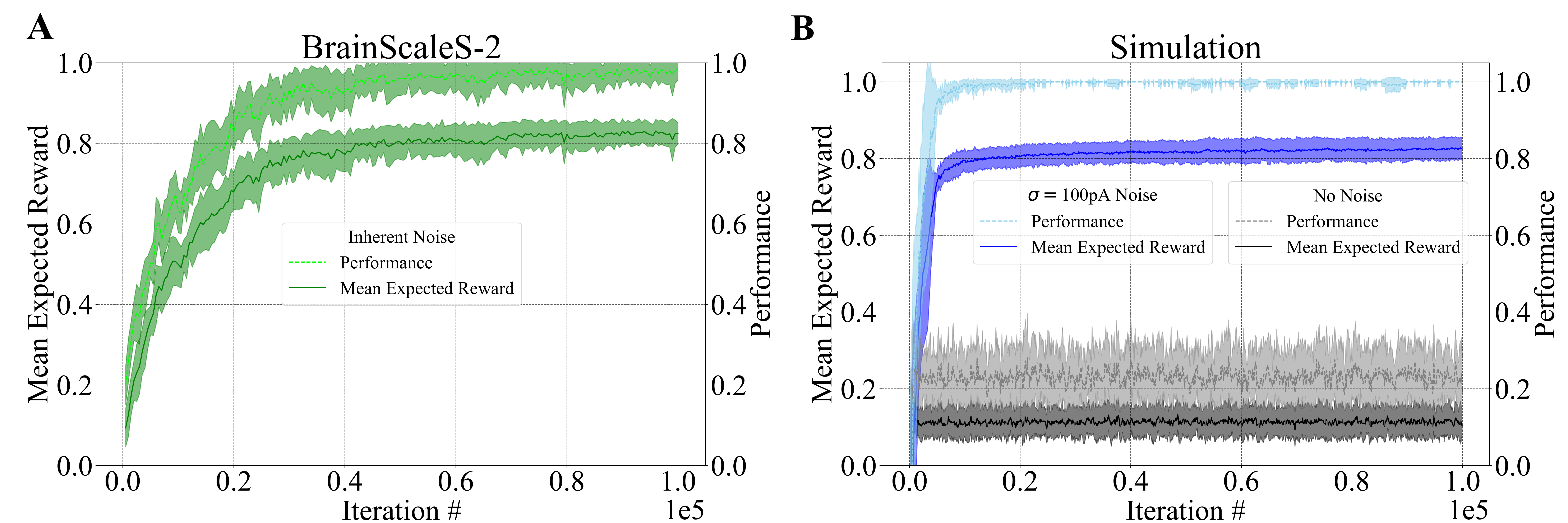}
    \caption{Learning results for BSS2 and the software simulation using NEST, in terms of Mean expected reward (\Cref{eq:mean-reward}) and Pong performance (\Cref{eq:performance}).
    In both cases, we plot the mean and standard deviation (shaded area) of 10 experiments.
    \textbf{A}: BSS2 uses its intrinsic noise as an action exploration mechanism that drives learning.
    \textbf{B}: The software simulation without noise is unable to learn and does not progress beyond chance level.
    Adding Gaussian zero-mean current noise with $\sigma=\SI{100}{\pico\ampere}$ to each neuron allows the network to explore actions and enables learning. 
    The simulation converges faster due to the idealized simulated scenario where no fixed-pattern noise is present.}
    \label{fig:learning}
\end{figure}

\subsection{Learning Performance}

The progress of learning over $10^5$ iterations is shown in \Cref{fig:learning} for both BSS2 (subplot A) and an ideal software simulation with and without injected noise (subplot B).
We use both measures described above to quantify the agent's success: the mean expected reward (\Cref{eq:mean-reward}) reflects the agent's aiming accuracy and the Pong performance (\Cref{eq:performance}) represents the ability of the agent to catch the ball using its elongated paddle.
By repeating the procedure with ten randomly initialized weight matrices, we show that learning is reproducible, with little variation in the overall progress and outcome.

\begin{figure}[h]
    \centering
    \includegraphics[width=15cm]{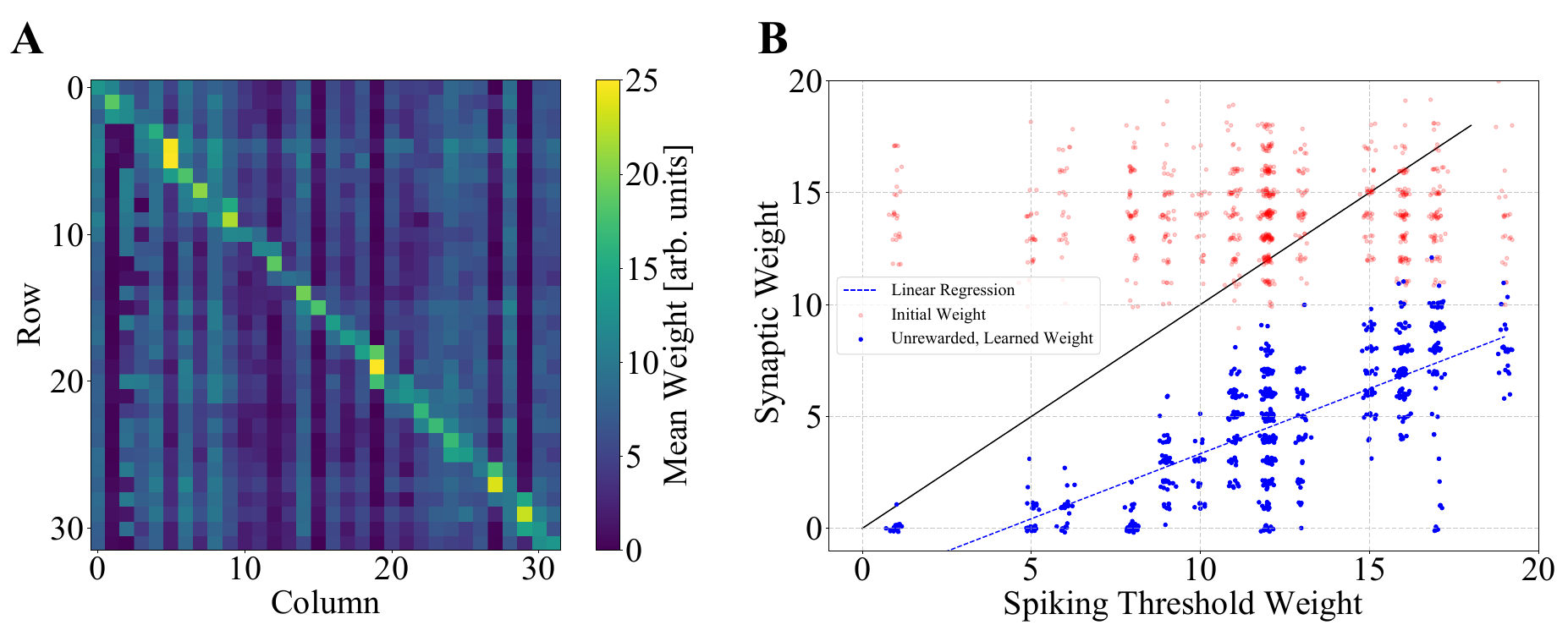}
    \caption{\textbf{A}: Synaptic weight matrix after learning, averaged over the $10$ trials depicted in \Cref{fig:learning}. 
    Weights on and near the diagonal dominate, corresponding to the goal of the learning task.
    The noticeable vertical stripes are a direct consequence of learning, which implicitly compensates neuronal variability (each column represents all input synapses of an action neuron). 
    \textbf{B}: Learning compensates circuit variability.
    Weights corresponding to unrewarded actions ($R=0$) are systematically pushed below the threshold weight of the respective neuron, i.e., below the main diagonal.
    This leads to a correlation of learned weight and threshold (Pearson's $r=0.76$).
    Weights are plotted with slight jitter along both axes for better visibility.}
    \label{fig:weights-and-comp}
\end{figure}

The optimal solution of the learning task is a one-to-one mapping of states to actions that place the center of the paddle directly below the ball at all times.
In terms of the neural network, this means that the randomly initialized weight matrix should be dominated by its diagonal elements.
We show the weight matrix on BSS2 after $10^5$ learning iterations, averaged over the ten different trials depicted in \Cref{fig:learning}, in \Cref{fig:weights-and-comp} A.
As expected, the diagonals of the matrix  are dominant.
Note that also slightly off-diagonal synapses are also strengthened, as dictated by the graded reward scheme.
The visibly distinguishable vertical lines stem from neuronal fixed-pattern noise, as learning adapts weights to compensate for neuronal variability and one column in the weight matrix corresponds to one neuron.

A screen recording of a live demonstration of the experiment is available at \url{https://www.youtube.com/watch?v=LW0Y5SSIQU4} and allows the viewer to follow the learning progress in a single experiment. 

\subsubsection{Temporal Variability on BSS2 Causes Exploration}

\label{sec:tempVariability}
On BSS2, action exploration and thereby learning is driven by trial-to-trial variations of neuronal spike counts that are due to temporal variability (see \Cref{fig:trial-to-trial} C, D).

In contrast, we find that the software simulation without injected noise (see \Cref{sec:sim}) is unable to progress beyond the mean expected reward received by an agent choosing random actions, which is around $\langle \bar R \rangle=0.1$ (the randomness in the weight matrix initialization leads to some variation in the mean expected reward after learning).
The only non-deterministic mechanism in the software simulation without noise is due to the fact that if several action neurons elicit the same number of spikes, the winner is chosen randomly among them, but its effects are negligible.
Injecting Gaussian current noise with zero mean and a standard deviation of $\sigma=\SI{100}{\pico\ampere}$ into each neuron independently enables enough action exploration to converge to similar performance as BSS2.
Compared to BSS2, the simulation with injected noise converges faster and to a higher level of Pong performance; this is due to the fact that the simulation contains no fixed-pattern noise, the network starts from an unbiased, perfectly balanced state and that Gaussian current noise is not an exact model of the temporal variability found in BSS2.

We can therefore conclude that under an appropriate learning paradigm, analog-hardware-specific temporal variability that is generally regarded as a nuisance can become a useful feature.
Adding an action exploration mechanism similar to $\epsilon$-greedy action selection would enable the software simulation to learn with guaranteed convergence to optimal performance, but would come at the cost of additional computational resources required for emulating such a mechanism.

\subsection{Learning Is Calibration}

\begin{figure}[h]
\begin{center}
\includegraphics[width=15cm]{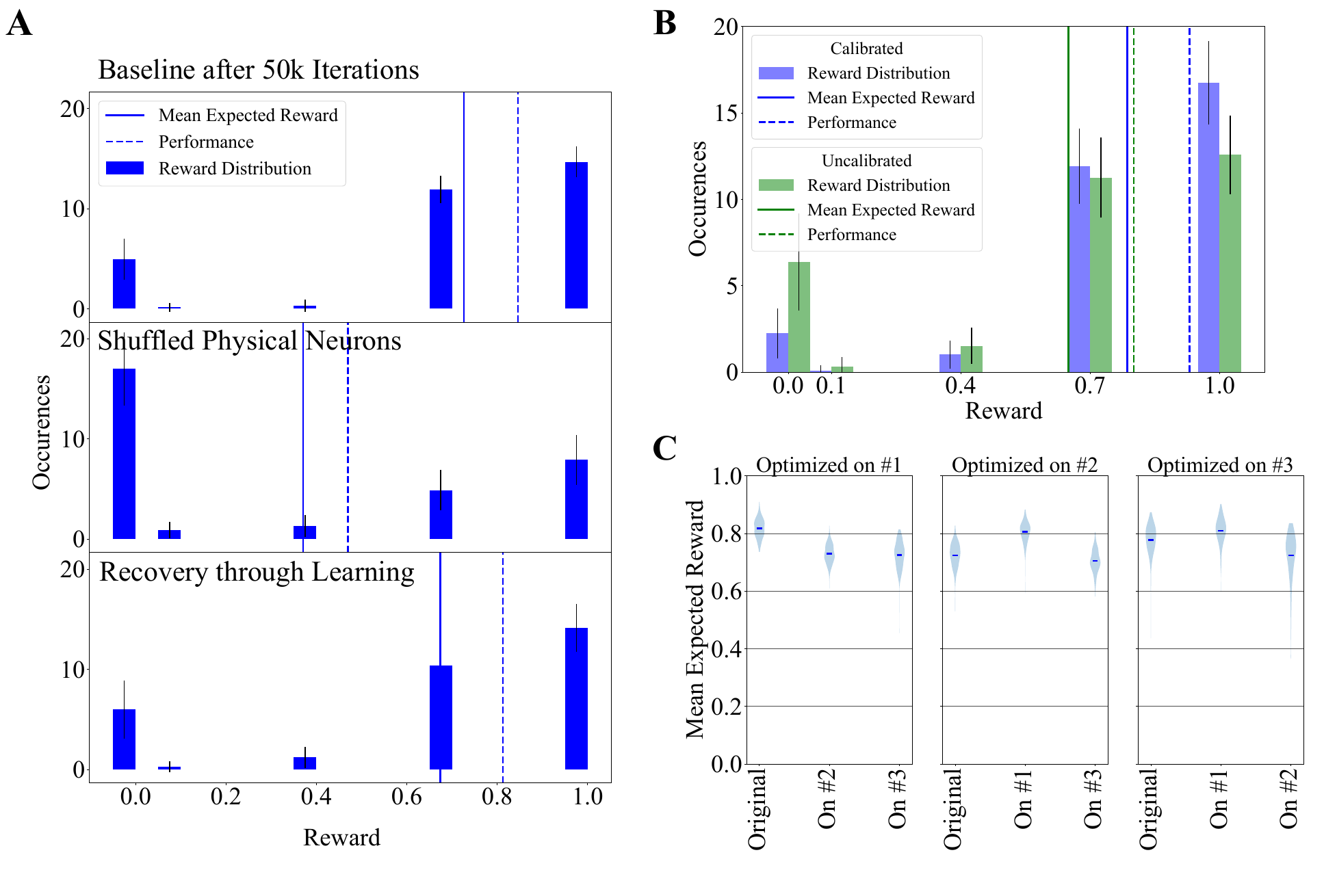}
\end{center}
\caption{\label{fig:robustness}
Learning can largely supplant individual calibration of neuron parameters and adapts synaptic weights to compensate for neuronal variability.
\textbf{A}: Top: Reward distribution over $100$ experiments measured with a previously learned, fixed weight matrix. 
Middle: Same as above, with randomly permuted neurons in each of the $100$ experiments. 
The agent's performance and therefore its received reward decline due to the weight matrix being adapted to a specific pattern of fixed-pattern noise.  
Bottom: Allowing the agent to learn for $50000$ additional iterations after having randomly shuffled its neurons leads to weight re-adaptation, increasing its performance and received reward.
In these experiments, LIF parameters were not calibrated individually per neuron. 
\textbf{B}: Reward distribution after $50000$ learning iterations in $100$ experiments for a calibrated and an uncalibrated system, with learning being largely able to compensate for the difference.
\textbf{C}: Results can be reproduced on different chips.
Violin plot of mean expected reward after hyperparameter optimization on chips \#1, \#2 and \#3.
Results are shown for the calibrated case.
All other results in this manuscript were obtained using Chip \#1.}
\end{figure}

The learning process adjusts synaptic weights such that individual differences in neuronal excitability on BSS2 are compensated.
We correlated learned weights to neuronal properties and found that learning shapes a weight matrix that is adapted to a specific pattern of neuronal variability.

Each neuron is the target of $32$ synapses.
A subset of these are systematically potentiated, as they correspond to actions yielding a reward higher than the current expected reward, while the remainder is depressed, as these synapses correspond to actions yielding less reward.
This leads to the diagonally-dominant matrix depicted in \Cref{fig:weights-and-comp} A. 
At the same time, neuronal variability leads to different spiking-threshold weights, i.e., the smallest synaptic weight for which the neuron elicits one spike  with a probability higher than \SI{5}{\percent}, given the fixed input spike train used in the experiment (see \Cref{fig:trial-to-trial} D).
The learning process pushes the weights of unrewarded ($R=0$) synapses below the spiking threshold of the respective neuron, thereby compensating variations of neuronal excitability, leading to a correlation of both quantities.
Using the weights depicted in (A) and empirically determined spiking thresholds, we found that for each neuron the weights of synapses which correspond to unrewarded actions were correlated with the spiking-threshold with a correlation coefficient of $r=0.76$  (see \Cref{fig:weights-and-comp} B).

The learned adaptation of the synaptic weights to neuronal variability can be disturbed by randomly shuffling the assignment of logical action units to physical neurons. 
This is equivalent to the thought experiment of physically shuffling neurons on the chip and leads to synaptic weights which are maladapted to their physical substrate, i.e., efferent neuronal properties.
In the following, we demonstrate the detrimental effects of such neuronal permutations and that the system can recover performance by subsequent learning.

We considered a weight matrix after $50000$ learning iterations and measured the resulting reward distribution over $100$ experiments with learning turned off.
Here, ``reward distribution'' refers to the distribution of the most recent reward over all $32$ states.
The top panel of \Cref{fig:robustness} A shows the reward distribution for this baseline measurement.
Mean expected reward and performance for this measurement are $\langle \bar R \rangle = \SI{0.73(05)}{}$ and $P= \SI{0.85(06)}{}$, respectively.

The same weight matrix was applied to $100$ systems with randomly shuffled physical neuron assignment and the reward distribution measured as before with learning switched off, yielding the distribution depicted in the middle panel of the same plot, with mean expected reward and performance of $\langle \bar R \rangle = \SI{0.37(09)}{}$ and $P=\SI{0.47(11)}{}$.
Each of the $100$ randomly shuffled systems was then subjected to $50000$ learning iterations, starting from the maladapted weight matrix, leading to the distribution shown in the bottom panel, with mean expected reward and performance of $\langle \bar R \rangle = \SI{0.67(07)}{}$ and $P=\SI{0.81(09)}{}$.

This demonstrates that our learning paradigm implicitly adapts synaptic weights to a specific pattern of neuronal variability and compensates for fixed-pattern noise. In a more general context, these findings support the idea that for neuromorphic systems endowed with synaptic plasticity, time-consuming and resource-intensive calibration of individual components can be, to a large extent, supplanted by on-chip learning.

\subsection{Learning Robustness}

One of the major concerns when using analog electronics is the control of fixed-pattern noise.
We can partly compensate for fixed-pattern noise using a calibration routine \citep{Aamir2018AnArchitecture}, but this procedure is subject to a trade-off between accuracy and the time and computational resources required for obtaining the calibration data.
Highly precise calibration is costly because it requires an exhaustive mapping of a high-dimensional parameter space, which has to be done for each chip individually.
A faster calibration routine, on the other hand, necessarily involves taking shortcuts, such as assuming independence between the influence of hardware parameters, thereby potentially leading to systematic deviations from the target behavior.
Furthermore, remaining variations can affect the the transfer of networks between chips and therefore potentially impact learning success when using a given set of (hyper-)parameters.
We discuss these issues in the following.
All results in this section were obtained after using $50000$ training iterations, which take around \SI{25}{\second} of wall-clock time on our BSS2 prototypes.

\subsubsection{Impact of Time Constant Calibration}
\label{sec:calib-robustness}

The neuronal calibration (see \Cref{sec:calib}) adjusts hardware parameters controlling the LIF time constants  ($\tau_\textrm{mem}$, $\tau_\textrm{ref}$ and $\tau_\textrm{syn}$) on a per-neuron basis to optimally match target time constants (\Cref{tab:parameters}), compensating neuronal variability. 
Depending on the target parameter value, it is possible to reduce the standard deviations of the LIF time constants across a chip by up to an order of magnitude \citep{Aamir2018AnArchitecture}.

We investigated the effect of uncalibrated neuronal time constants on learning.
The uncalibrated state is defined by using the same value for a given hardware parameter for all neurons on a chip.
To set up a reasonable working point, we chose this to be the average of the calibrated values of all neurons on the chip.
A histogram of the membrane time constants of all neurons on the chip in the calibrated and uncalibrated state is given in \Cref{fig:trial-to-trial} B.
In both cases, voltages ($v_\textrm{leak}$, $v_\textrm{thresh}$, $v_\textrm{reset}$) were uncalibrated.

We measured the reward distribution after learning in both the calibrated and the uncalibrated state, performing $100$ experiments in both cases (\Cref{fig:robustness} B). Even in the uncalibrated state, learning was possible using only the inherent hardware noise for action exploration, albeit with some loss (around \SI{17}{\percent}) in mean expected reward.
The mean expected reward and performance in the calibrated state are $\langle \bar R \rangle = \SI{0.79(05)}{}$ and $P=\SI{0.93(05)}{}$, respectively. 
In the uncalibrated state, the values are $\langle \bar R \rangle = \SI{0.65(08)}{}$ and $P=\SI{0.80(09)}{}$.

As a corollary, the reward distributions depicted in \Cref{fig:robustness} B suggest that narrowing the reward window (defined in \Cref{eq:reward}) to half the original size (i.e.,\ removing the $R\leq0.4$ part) would only have a small effect on converged Pong performance.

\subsubsection{Transferability of Results Between Chips}
\label{sec:transfer}

All results presented thus far were obtained using one specific chip (henceforth called chip \#1) and parameter set (given in \Cref{tab:parameters} as set \#1).
These parameters were found using a hyper-parameter optimization procedure (see \Cref{sec:mandm}) that was performed on this chip.
In addition, we performed the same optimization procedure on two other chips, using the original optimization results as a starting point, which yielded three parameter sets in total.
The two additional parameter sets, sets \#2 and \#3 corresponding to chips \#2 and \#3, are also given in \Cref{tab:parameters}.

We then investigated the effects of transferring each parameter set to the other chips, testing all six possible transitions of learned parameters between the chips.
To this end, we conducted $200$ trials consisting of $50000$ learning iterations on every chip for every parameter set and compare the resulting mean expected reward  in \Cref{fig:robustness} C.
Learning results were similar across all nine experiments.
As expected due to process variations, there were small systematic deviations between the chips, with chip \#1 slightly outperforming the other two for all three hyperparameter sets, but in all scenarios the agent successfully learned to play the game.
These results suggest that chips can be used as drop-in replacements for other chips and that the hyperparameter optimization does not have to be specific to a particular substrate to yield good results.

\subsection{Speed and Power Consumption}

\begin{figure}[h]
    \centering
    \includegraphics[width=10cm]{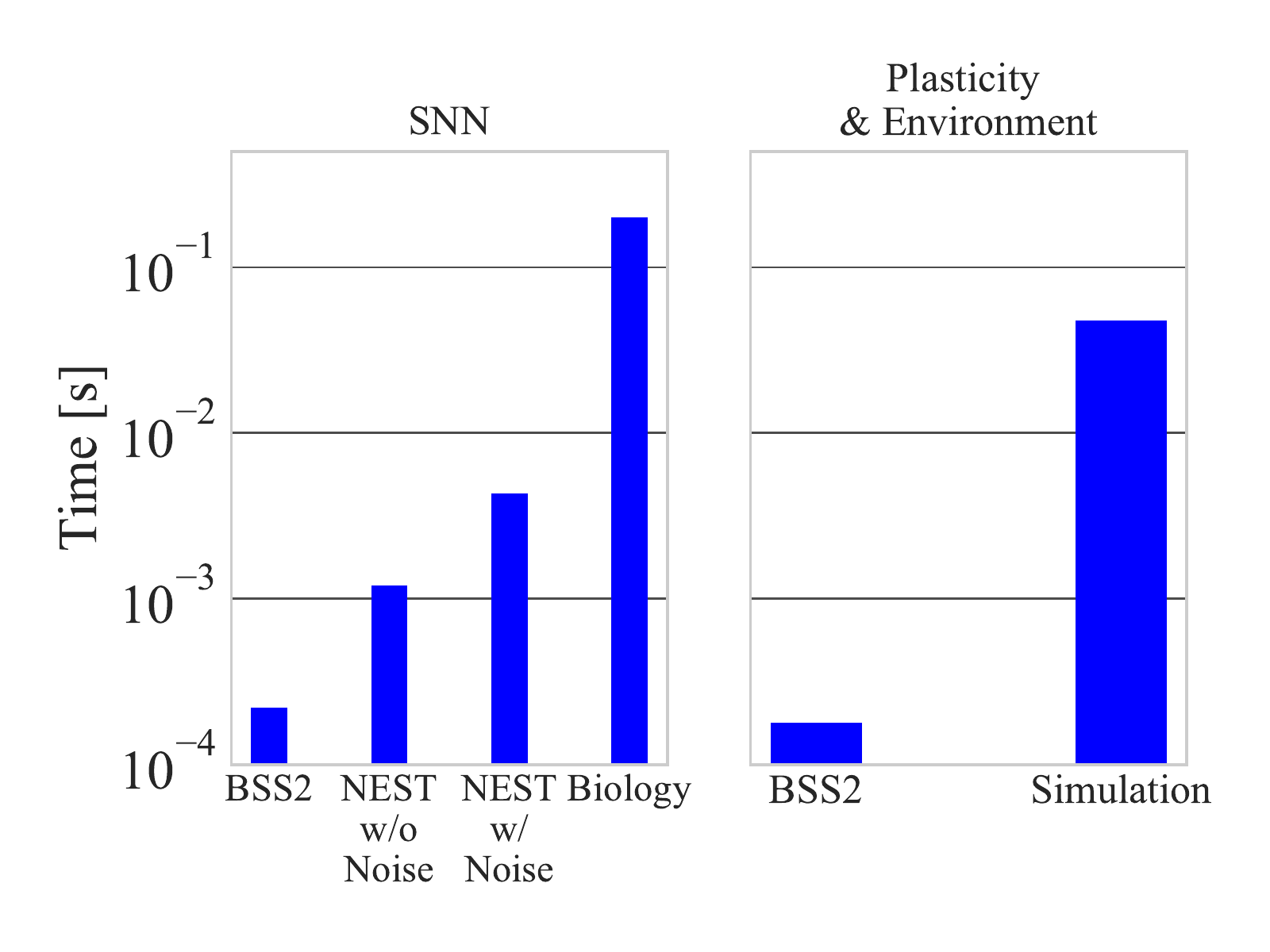}
    \caption{A single experiment iteration on BSS2 takes around \SI{400}{\micro\second}, with \SI{220}{\micro\second} devoted to network emulation and \SI{180}{\micro\second} to plasticity calculation. 
    The spiking network activity corresponds to 200 biological ms.
    In the software simulation, the neural network with and without noise is simulated in \SI{4.3}{\milli\second} and  \SI{1.2}{\milli\second}, respectively, while plasticity calculations take around \SI{50}{\milli\second}.}
    \label{fig:time}
\end{figure}

\label{sec:speed-power}

\subsubsection{Speed}
\label{sec:speed}
An essential feature of BSS2 is its $10^3$ speed-up factor compared to biological real-time.
To put this in perspective, we compared the emulation on our BSS2 prototypes to a software simulation with NEST running on a conventional CPU (see \Cref{sec:sim}).
We found that a single experiment iteration in the simulation takes approximately \SI{50}{\milli\second} when running it on a single core of an Intel i7-4771 CPU (utilizing more cores does not lead to a faster simulation due to the small size of the network).
For our speed comparison, we differentiate the case of no injected noise (where the network's time evolution is fully deterministic) and the case where we use NEST's \textit{noise\_generator} module to inject current noise into each neuron.
When noise is injected, the \SI{200}{\milli\second} of neural network activity is simulated in \SI{4.3}{\milli\second}; without noise, the activity is simulated in \SI{1.2}{\milli\second}.
This is the time that is spent in NEST's state propagation routine (the \textit{Simulate} routine).
The remainder is spent calculating the plasticity rule and environment in the additional Python script, which we consider separately as it was external to NEST.

In contrast, BSS2 takes \SI{0.4}{\milli\second} per iteration as measured using the time stamp register of the PPU.
This time is approximately equally divided between neural network emulation and other calculations, including the updates of the synaptic weights and the environment.
The total time duration of an experiment with $50000$ learning iterations is \SI{25}{\second} for BSS2 and \SI{40}{\minute} for the software simulation. 
A constant overhead of around \SI{5}{\second} when using the chip is spent on calibration, connection setup and configuring the chip.

The comparison between BSS2 and the software simulation is visualized in \Cref{fig:time}.
Note that while the calculation of eligibility traces in software was not optimized for speed, it incurs a significant computational cost, especially as it scales linearly with the number of synapses and therefore approximately quadratically with the network size. 
This is in contrast to the emulation on BSS2, where the eligibility trace is measured locally with analog circuitry at each synapse during network emulation.

In both cases, the time taken for environment simulation is negligible compared to plasticity calculation. 
We have confirmed that these measurements are independent on learning progress (i.e., measurements during the first iterations and with a diagonal weight matrix are equal). 

\subsubsection{Power Consumption}
\label{sec:power}
We measured the current drawn by the CPU during the software simulation of the neural network, i.e., during NEST's numerical simulation routine as well as when idling using the EPS \SI{12}{\volt} power supply cable that supplies only the CPU.
Again, we differentiate the cases of no noise and injected current noise.
Based on these measurements, we determined a lower bound of \SI{24}{\watt} for the CPU power consumption during SNN simulation without noise and a lower bound of \SI{25}{\watt} when injecting noise.
A single simulation iteration in software without noise, taking \SI{1.2}{\milli\second} and excluding plasticity and environment simulation, therefore consumes at least \SI{29}{\milli\joule}.
When we inject current noise, the simulation takes \SI{4.3}{\milli\second} which corresponds to an energy consumption of \SI{106}{\milli\joule} per iteration.

By measuring the current drawn by BSS2 during the experiment, we calculated the power dissipated by it to be \SI{57}{\milli\watt}, consistent with the measurement of another network in \citep{Aamir2018AnArchitecture} (this is not considering the power drawn by the FPGA, which is only used for initial configuration, and other prototype board components). 
This implies a per-iteration energy consumption, including plasticity and environment simulation, of around \SI{23}{\micro\joule} by the chip.

These measurements imply that, in a conservative comparison, the emulation using BSS2 is at least three orders of magnitude more energy-efficient than the software simulation.

\section{Discussion and Outlook}
We demonstrated key advantages of our approach to neuromorphic computing in terms of speed and energy efficiency compared to a conventional approach using dedicated software.
The already observable order-of-magnitude speed-up of our BSS2 prototypes compared to software for our small-scale test case is expected to increase significantly for larger networks.
At the same time, this performance can be achieved with a three-orders-of-magnitude advantage in terms of power consumption, as our conservative estimate shows.

Furthermore, we showed how substrate imperfections, which inevitably lead to parameter variations in analog neuro-synaptic circuits, can be compensated using an implementation of reinforcement learning via neuromodulated plasticity based on R-STDP.
Meta-learning can be done efficiently despite substrate variability, as results of hyperparameter optimization can be transferred across chips.
We further find that learning is not only robust against analog temporal variability, but that trial-to-trial variations are in fact used as a computational resource for action exploration.

In this context, it is important to mention that temporal variability is generally undesirable due to its uncontrollable nature and chips are designed with the intention of keeping it to a minimum.
Still, learning paradigms that handle such variability gracefully are a natural fit for analog neuromorphic hardware, where noise inevitably plays a role.

Due to the limited size of the chips used in this pilot study, the neural networks discussed here are constrained in size.
However, the final system will represent a significant scale-up of this prototype.
The next hardware revision will be the first full-size BSS2 chip and will provide 512 neurons, 130k synapses and two embedded processors with appropriately scaled vector units.
Neurons on the chip will emulate the Adaptive-Exponential (AdEx) Integrate-and-Fire model with multiple compartments, while synapses will contain additional features for Short-Term Plasticity (STP).
Poisson-like input stimuli with rates in the order of \SI{}{\mega\hertz}, i.e., biological \SI{}{\kilo\hertz}, will be realized using pseudo-random number generators to provide the possibility for stochastic spike input to neurons, yielding controllable noise which can be used for emulation of in-vivo activity and large-scale functional neural networks \citep{Destexhe2003TheVivo,petrovici2014characterization,Petrovici2016StochasticState}.
The full BSS2 neuromorphic system will comprise hundreds of such chips on a single wafer which itself will be interconnected to other wafers, similar to its predecessor BrainScaleS~1 \citep{Schemmel2010AModeling}.
The study at hand lays the groundwork for future experiments with reinforcement learning in neuromorphic SNNs, where an expanded hardware real-estate will allow the emulation of more complex agents learning to navigate more difficult environments.

The advantages in speed and energy consumption will become even more significant when moving to large networks.
In our experiments, the relative speed of the software simulation was due to the small size of the network.
State-of-the-art software simulations of large LIF neural networks take minutes to simulate a biological second \citep{Jordan2018ExtremelyComputers} and even digital simulations on specialized neuromorphic hardware typically achieve real-time operation \citep{vanAlbada2018PerformanceModel,mikaitis2018_neuromodulated}.
On BSS2, the speed-up factor of \num{e3} is independent of network size, which, combined with the two embedded processors per chip and the dedicated synapse circuits containing local correlation sensors, enables the accelerated emulation of large-scale plastic neural networks.

The speed-up factor of BSS2 is both an opportunity and a challenge.
In general, rate-based or sampling-based codes would profit from longer integration times enabled by the acceleration.
On the other hand, interfacing BSS2 to the real world (e.g.\ using robotics) requires fast sensors; the speed-up could enable fast sensor-motor loops with possible applications in, for example, radar beam shaping.
In general, however, the main advantage of an accelerated system becomes most evident when learning is involved: long-term learning processes lasting several years in biological spiking networks can be emulated in mere hours on BSS2, whereas real-time simulations are unfeasible. 

The implemented learning paradigm (R-STDP) and simulated environment (Pong) were kept simple in order to focus our study on the properties of the prototype hardware.
R-STDP is a well-studied model that lends itself well to hardware implementation, while the simplified Pong game is a suitable learning task that can be realized on the prototype chip and provides an accessible, intuitive interpretation.
Still, the PPU's computational power limits the complexity of environment simulations that can be realized on-chip, especially when simulations have hard real-time constraints.
However, such simulations could take place on a separate system that merely provides spike input, collects spike output and provides reward to the neuromorphic system.
Alternatively, simulations could be horizontally distributed across PPUs.

Solving more complex learning tasks demands more complex network models.
We expect that the future large-scale BSS2 system will be able to instantiate not only larger, but also more complex network models, by offering more flexibility in the choice of spiking neuron dynamics (e.g.\ AdEx, LIF), short-term synaptic plasticity and enhanced PPU capabilities.
However, further theoretical work is required for mapping certain state-of-the-art reinforcement learning models, such as DQN \citep{Mnih2015Human-levelLearning} and  AlphaGo (Zero) \citep{Silver2016MasteringSearch,Silver2017MasteringKnowledge} to this substrate.
On the other hand, learning paradigms like TD-STDP \citep{Fremaux2013ReinforcementNeurons}, which implements actor-critic reinforcement learning using a spiking neural network, already match the capabilities of a large-scale version of the BSS2 system and therefore represent suitable candidates for learning in more complex environments with sparse rewards and agent-dependent state transitions.

\section*{Conflict of Interest Statement}

The authors declare that the research was conducted in the absence of any commercial or financial relationships that could be construed as a potential conflict of interest.

\section*{Author Contributions}
Andreas Hartel (AH), Akos F. Kungl (AFK), Timo Wunderlich (TW) and Mihai A. Petrovici (MAP) designed the study.
TW conducted the experiments and the evaluations.
TW and AFK wrote the initial manuscript.
Eric Müller (ECM) and Christian Mauch (CM) supported experiment realization; ECM coordinated the software development for the neuromorphic systems.
Yannik Stradmann (YS) contributed with characterization, calibration testing and debugging of the prototype system.
Syed Ahmed Aamir (SAA) designed the neuron circuits.
Andreas Grübl (AG) was responsible for chip assembly and did the digital front- and backend implementation.
Arthur Heimbrecht (AWH) extended the GNU Compiler Collection backend support for the embedded processor.
David Stöckel (DS) contributed to the host control software development and supported chip commissioning.
Korbinian Schreiber (KS) designed and assembled the development board.
Christian Pehle (CP) provided FPGA firmware and supported chip commissioning.
Gerd Kiene (GK) contributed to the verification of the synaptic input circuits.
Johannes Schemmel (JS) is the architect and lead designer of the neuromorphic platform.
MAP, Karlheinz Meier (KM), JS, Sebastian Billaudelle (SB), and ECM provided conceptual and scientific advice.
All authors contributed to the final manuscript.

\section*{Funding}
This research was supported by the EU 7th Framework Program under grant agreements 269921 (BrainScaleS), 243914 (Brain-i-Nets), 604102 (Human Brain Project), the Horizon 2020 Framework Program under grant agreement 720270 (Human Brain Project), the Manfred Stärk Foundation and the Deutsche Forschungsgemeinschaft within the funding programme Open Access Publishing, by the Baden-Württemberg Ministry of Science, Research and the Arts and by Ruprecht-Karls-Universität Heidelberg.

\section*{Acknowledgments}
The authors wish to thank Simon Friedmann, Matthias Hock and Paul Müller.
They contributed to developing the hardware and software that enabled this experiment.

\bibliographystyle{unsrtnat}
\begin{singlespace}
\bibliography{mendeley_v2}

\pagebreak
\section*{Appendix}

\begin{table}[h]
\begin{tabular}{lllll}
Symbol & Description & Value \\
\toprule
& Neuromorphic hardware & BrainScaleS 2 (2nd prototype version) \\
$N$ & Number of action/output neurons (LIF)& $32$ \\
$N_\textrm{S}$ & Number of state/input units & $32$ \\
$N_\textrm{syn}$ & Number of synapses & $32\cdot32=1024$ \\
$N_\textrm{spikes}$ & Number of spikes from input unit & $20$ \\
$T_\textrm{ISI}$ & ISI of spikes from input unit & \SI{10}{\micro\second} \\
$w$ & Mean of distribution of initial weights (digital value)& $14$ \\
$\sigma_w$ & Standard deviation of distribution of initial weights & $2$ \\
$L$ & Length and width of quadratic playing field & 1 \\
$\|\vec v \|_1$ & L1-norm of ball velocity & $0.025$ per iteration \\
$v_\textrm{p}$ & Velocity of paddle controlled by BSS2 & $0.05$ per iteration \\
$r_\textrm{b}$ & Radius of ball & $0.02$ \\
$r_\textrm{p}$ & Length of paddle & $0.20$ \\
$\gamma$ & Decay constant of reward  & $0.5$ \\
$\beta$ & Learning rate & $0.125$ \\
&NEST version (software simulation) & 2.14.0 \\
&NEST timestep                      & \SI{0.1}{\milli\second} \\
&CPU (software simulation, one core used) & Intel i7-4771 \\
\end{tabular}
\begin{tabular}{lllll}
\\ % empty line
 && Set \#1  & Set \#2  & Set \#3  \\
 && (standard) & & \\
\toprule 
$\tau_\textrm{mem}$ & LIF membrane time constant & \SI{28.5}{\micro\second} & \SI{18.4}{\micro\second} & \SI{24.8}{\micro\second} \\
$\tau_\textrm{ref}$ & LIF refractory time constant & \SI{4}{\micro\second} & \SI{14.3}{\micro\second} & \SI{13.8}{\micro\second} \\
$\tau_\textrm{syn}$ & LIF excitatory synaptic time constant & \SI{1.8}{\micro\second} & \SI{2.4}{\micro\second} & \SI{1.4}{\micro\second} \\
$v_\textrm{leak}$ & LIF leak voltage & \SI{0.62}{\volt} & \SI{0.56}{\volt} & \SI{0.87}{\volt} \\
$v_\textrm{reset}$ & LIF reset voltage & \SI{0.36}{\volt} & \SI{0.36}{\volt} & \SI{0.30}{\volt} \\
$v_\textrm{thresh}$ & LIF threshold voltage & \SI{1.28}{\volt} & \SI{1.31}{\volt} & \SI{1.21}{\volt} \\
$\eta_+$ & Amplitude of correlation function $a_+$ (digital value) & $72$ & $114$ & $70$ \\
$\tau_+$ & Time constant of correlation function $a_+$ & \SI{64}{\micro\second} & \SI{80}{\micro\second} & \SI{60}{\micro\second} \\
\end{tabular}\caption{\label{tab:parameters}
Parameters used in the experiment. The different parameter sets are the result of optimizing parameters on three different chips. If not mentioned otherwise, results were obtained using set \#1. Abbreviations used in table: LIF: Leaky Integrate-and-Fire; ISI: Inter-Spike Interval.
}
\end{table}
\end{singlespace}

\end{document}